\def\year{2020}\relax
\documentclass[letterpaper]{article} 
\usepackage{aaai20}  
\usepackage{times}  
\usepackage{helvet} 
\usepackage{courier}  
\usepackage[hyphens]{url}  
\usepackage{graphicx} 
\usepackage{graphicx}  
\usepackage{amsmath}
\usepackage{amssymb}
\usepackage{color}
\usepackage{enumitem}

\usepackage{algorithm}
\usepackage[noend]{algpseudocode}

\usepackage{amsmath}
\usepackage{bm}
\usepackage{multirow}
\usepackage{makecell}

\usepackage{mathtools}
\usepackage{dsfont}
\usepackage{multirow}
\usepackage{tabularx}
\usepackage{arydshln}

\newcommand{\defeq}{\vcentcolon=}
\newcommand{\eqdef}{=\vcentcolon}

\newcommand{\hyeonwoo}[1]{{\color{red}{HW: #1}}}
\newcommand{\ilchae}[1]{{\color{blue}{IC: #1}}}

\def \x{\mathbf{x}}
\def \f{\mathbf{f}}
\def \y{\mathbf{y}}

\def \datainit{\mathcal{D}_\text{init}}
\def \dataonline{\mathcal{D}_\text{on}}
\def \dataval{\mathcal{D}_\text{test}}
\def \phigeneral{\boldsymbol{\phi}}
\def \thetageneral{\boldsymbol{\theta}}
\def \thetainit{{\boldsymbol{\theta}_\text{init}^0}}
\def \thetainitadapt{\boldsymbol{\theta}_\text{init}}
\def \thetaonlineadapt{\boldsymbol{\theta}_\text{on}}
\def \alphainit{\boldsymbol{\alpha}_\text{init}}
\def \alphaonline{\boldsymbol{\alpha}_\text{on}}
\def \Ainit{\mathcal{A}_\text{init}}
\def \Aonline{\mathcal{A}_\text{on}}
\def \Kinit{K_\text{init}}
\def \Konline{K_\text{on}}
\def \lossval{\mathcal{L}_\text{test}}
\def \losstriple{\mathcal{L}_\text{tri}}
\def \lossmeta{\mathcal{L}_\text{meta}}
\def \losslasso{\mathcal{L}_\text{lasso}}
\def \lossepisode{\mathcal{L}_\text{ep}}
\def \lossprune{\mathcal{L}_\text{prune}}

\title{Real-Time Object Tracking via Meta-Learning: \\ Efficient Model Adaptation and One-Shot Channel Pruning}

\author{Ilchae Jung$^{1,2}$ \hspace{0.5cm} Kihyun You$^1$ \hspace{0.5cm} Hyeonwoo Noh$^{1,2}$ \hspace{0.5cm} Minsu Cho$^1$ \hspace{0.5cm} Bohyung Han$^2$ \\
	$^1$Computer Vision Lab., POSTECH, Korea \\ 
	$^2$Computer Vision Lab. ECE, \& ASRI, Seoul National University, Korea   \\ 
	 \null \{chey0313, kihyun13, shgusdngogo, mscho\}@postech.ac.kr \hspace{1cm}  bhhan@snu.ac.kr
  \null
}

\begin{document}
\maketitle
\begin{abstract}

      We propose a novel meta-learning framework for real-time object tracking with efficient model adaptation and channel pruning. 
Given an object tracker, our framework learns to fine-tune its model parameters in only a few gradient-descent iterations during tracking while pruning its network channels using the target ground-truth at the first frame. 
Such a learning problem is formulated as a meta-learning task, where a meta-tracker is trained by updating its meta-parameters for initial weights, learning rates, and pruning masks through carefully designed tracking simulations.  
The integrated meta-tracker greatly improves tracking performance by accelerating the convergence of online learning and reducing the cost of feature computation. 
Experimental evaluation on the standard datasets demonstrates its outstanding accuracy and speed compared to the state-of-the-art methods.

\end{abstract}

\section{Introduction}
\label{sec:introduction}
Recent advances in deep neural networks have drastically improved visual object tracking technology. 
By learning strong representations of target and background using convolutional neural networks, many algorithms~\cite{nam2016learningmulti,nam2016modeling,han2017branchout,martin2016beyond,martin2017eco} have achieved a significant performance gain.
Based on the success, recent methods~\cite{danelljan2019atom,li2019siamrpn++,huang2017learning,fan2017parallel,zheng2018distractor,jung2018realtime} have further advanced network architectures and training techniques for better accuracy and speed.
Such manual design choices, however, often result a suboptimal solution in a limited exploration space eventually.

Meta-learning~\cite{santoro2016meta,andrychowicz2016learning,finn2017model_agnostic} automates the optimization procedure of a learning problem through the evaluation over a large number of episodes.
It facilitates exploring a large hypothesis space and fitting a learning algorithm for a particular set of tasks.
Considering that an object tracking algorithm aims to learn a parameterized target appearance model specialized for individual video sequences, it is natural to adopt meta-learning for the effective optimization in tracking algorithms.
In the context of object tracking based on deep neural networks, an episode corresponds to a realization of object tracking based on a parameterized model for a video sequence.
The execution of an episode is computationally expensive and time-consuming, which makes it difficult to incorporate meta-learning into object tracking because it requires optimization over a large number of learning episodes.
Meta-Tracker~\cite{park2018metatracker} circumvents this issue by simulating a tracking episode within a single frame.
However, in this approach, meta-learning is limited to initializing the deep neural network for target appearance modeling at the first frame, where ground-truth target annotation is available.
This is because Meta-Tracker relies only on the accurate ground-truth for meta-learning, although tracking after the first frame involves model adaptation based on estimated targets from the previous frames.

We introduce a novel meta-learning framework for object tracking, which focuses on fast model adaptation.
In particular, our approach simulates the model adaptation by dividing it into the following two distinct cases: (1) initial adaptation, where the model is optimized for the one-shot target ground-truth at the first frame and (2) online adaptation, where the model is updated using the tracked targets in previous frames.
Such a fine-grained simulation enables to capture distinct properties of two cases and allows the meta-learned model to generalize better.
In our meta-learning for object tracking, we evaluate the learning algorithm in terms of its expected accuracy over various situations by simulating diverse challenging scenarios with hard examples. 
Moreover, we develop a one-shot channel pruning technique via meta-learning based on a single ground-truth target annotation at the first frame. 
%
Our main contributions are threefold:
\begin{itemize}[label=$\bullet$]
  \item We propose a meta-learning framework for object tracking, which efficiently optimizes network parameters for target appearance through sophisticated simulation and test of tracking episodes in meta-training.
  \item We introduce a one-shot network pruning technique via meta-learning, which enables to learn target-specific model compression based on only a single ground-truth annotation available at the first frame.
  \item We demonstrate that our meta-tracker leads to significant speed-up and competitive accuracy on the standard benchmark datasets.
\end{itemize}
%

\section{Related Work}
\label{sec:related}

Most of the state-of-the-art visual tracking algorithms rely on the representations learned from deep neural networks, and often formulate object tracking as a classification~\cite{nam2016learningmulti,jung2018realtime} or a metric learning problem~\cite{luca2016fullycovolutional,zheng2018distractor}.
While several recent algorithms show strong representation as well as computational efficiency, they still suffer from target appearance variations in challenging situations during tracking.

A pretraining stage is commonly adopted for a tracker to learn discriminative features for object tracking.
MDNet and its variants~\cite{nam2016learningmulti,jung2018realtime} employ multi-domain learning to simulate various tracking scenarios and learn generic target representations.
Meanwhile, \cite{luca2016fullycovolutional,jack2017endtoend,zheng2018distractor} discuss representation learning based on correlation filters.
However, these methods mainly focus on learning representations for target appearances, but the efficient optimization of model update procedure has been rarely explored.

Meta-learning is a framework to learn a learning algorithm under a certain distribution~\cite{thrun1998learning,hochreiter2001learning}.
It has been explored to expedite learning procedures in few-shot classification~\cite{santoro2016meta,andrychowicz2016learning,finn2017model_agnostic,li2017metaSGD}, reinforcement learning~\cite{finn2017model_agnostic,al2018continuous} and imitation learning~\cite{duan2017one}.
Meta-Tracker~\cite{park2018metatracker} adopts this idea for object tracking to demonstrate potential benefit for fast model updates.
However, it applies meta-learning only to the adaptation at the first frame of an input video, and fails to show the practical efficiency of tracking algorithms through fast online model updates.
Moreover, it relies on various heuristics to establish meta-learning, {\it e.g.,} learning rate adjustments, layer selections for parameter updates, and label shuffling.

Model compression is a useful technique to reduce the size of a deep neural network and accelerate its inference time~\cite{han2015learningboth,han16deep}.
A popular approach to model compression is channel pruning~\cite{wen2016learningsparsity,he2017channel}, which aims to remove a subset of channels in each layer based on their usefulness.
Channel pruning often involves a complex optimization task or time-consuming validation and fine-tuning, which is not affordable in real-time object tracking.
To tackle the limitation, \cite{choi2018context} employ multiple expert autoencoders to compress deep features for object tracking, where the targets are divided into several coarse categories and then allocated to autoencoders for feature compression according to their class labels.
However, the predefined categories limit the generality of the tracker and the multiple autoencoders are computationally expensive to learn.

Improving the efficiency of tracking algorithms by fast model adaptation and compression via meta-learning has not been explored.
The proposed meta-learning approach resembles~\cite{finn2017model_agnostic,li2017metaSGD} in the sense that we learn initial parameters and learning rates through meta-learning while we propose a novel method to estimate the distribution of the target task for stable meta-learning.
Contrary to \cite{park2018metatracker}, our algorithm performs more sophisticated parameter estimation without using the complex heuristics for training.
Also, our channel pruning is based on a one-shot ground-truth available at the first frame, and is realized within a meta-learning framework for efficient optimization.

\begin{figure*}[t]
	\begin{center}
		\includegraphics[width=\linewidth]{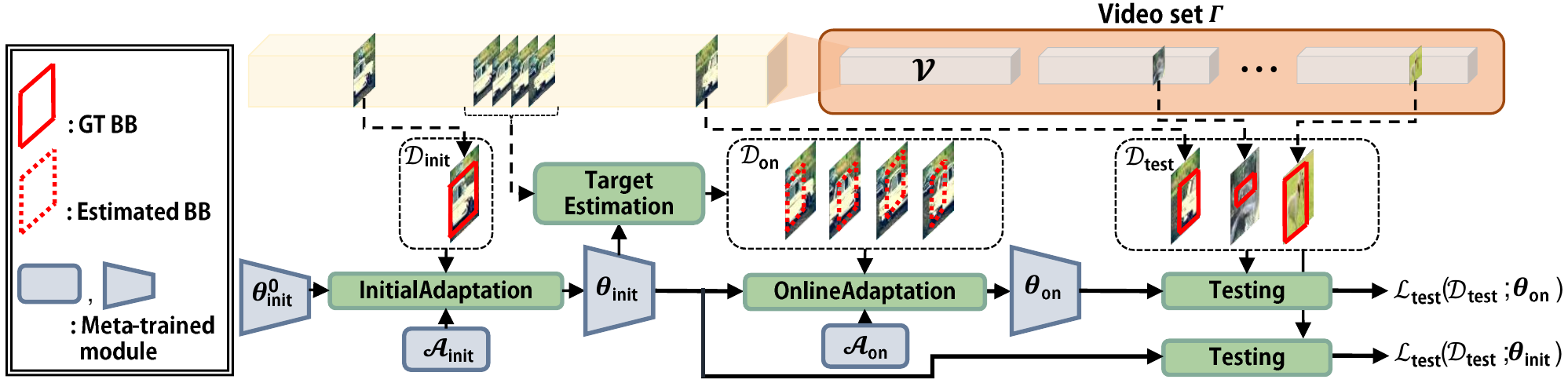}%
	\end{center}
	\caption{\textbf{A simulated tracking episode.} We meta-learn a fast model adaptation algorithm based on simulated episodes.  The model is adapted with ground-truth target $\datainit$ in the initial adaptation and with estimated target $\dataonline$ in the online adaptation. The meta-parameters ($\thetainit, \Ainit, \Aonline$) are learned to minimize test loss $\lossval$ in meta-training over multiple simulated episodes.} $\lossval$ is defined using $\dataval$ containing both ground-truth from upcoming frames and hard examples from other annotated videos. The hard examples are used to simulate diverse clutter appearing in the standard object tracking.
	
	\label{fig:overall_pipeline}
\end{figure*}

\section{Meta-learning for Fast Adaptation}
\label{sec:meta_learn_fast_adaptation}

This section presents an iterative framework for optimizing hyperparameters via meta-learning for fast model adaptation during object tracking. Building on MAML~\cite{finn2017model_agnostic},  our meta-learning framework is adapted for object tracking and thus consists of a series of tracking episode simulation, model test, and meta-parameter optimization.


\subsection{Objective}


The objective of model adaptation is to minimize the standard cross-entropy loss of an output $\f(\x;\thetageneral)\in\mathbb{R}^2$ from a model parametrized by $\thetageneral\in\mathbb{R}^d$, given an input patch $\x$ with its binary label $\y \in \{[1, 0]^\top, [0, 1]^\top\}$, where $[1, 0]^\top$ denotes the positive label while $[0, 1]^\top$ is the negative one.
Then, the loss function is formally given by
\begin{equation}
\mathcal{L}(\mathcal{D};\thetageneral)= -\mathbb{E}_{p_{\mathcal{D}}(\x, \y)} \Big[  \y^\top \log \big[\text{Softmax}\big(\f(\x;\thetageneral)\big)\big] \Big],
\label{eq:adaptation}
\end{equation}
where $\mathcal{D}=\{(\x_i, \y_i)\}_{i=1}^N$ is a dataset for learning target appearance, which is collected from tracking results in previously frames, and $p_{\mathcal{D}}(\x, \y)$ is the distribution of data in $\mathcal{D}$.
The softmax activation function $\text{Softmax}(\cdot)$ normalizes the output into a probability-like vector.

\subsection{Tracking Simulation}

Since the hyperparameters are meta-learned based on tracking simulation, the construction of realistic simulations at training time is crucial to the generalization of the meta-learned algorithms.
The typical object tracking algorithms consider two types of model adaptation~\cite{nam2016learningmulti,jung2018realtime}.
One is performed at the $1^\mathrm{st}$ frame using the ground-truth annotation while the other is at subsequent frames using estimated targets in previous frames.
We call the former {\em initial adaptation} and the latter {\em online adaptation}. The major difference between the initial adaptation and the online adaptation stems from the availability of ground-truth labels in learning. 
We consider this aspect in the tracking simulation and use a different dataset for each adaptation during meta-learning.

Let $\Gamma$ be a training video set and $\mathcal{V} \in \Gamma$ is an annotated video with tracking ground-truths for a single moving target.
Specifically, an annotated video $\mathcal{V}$ is a sequence of tuples consisting of a frame and a ground-truth target annotation for the frame.
We define the \textit{tracking simulation} as a sequence of initial adaptation, online adaptation, and test dataset construction in meta-training for a target annotated in $\mathcal{V}$.
Unlike the standard object tracking, which aims to estimate targets in every frame, tracking simulation performs only a single initial adaptation and a single online adaptation on the datasets that are artificially generated from $\mathcal{V}$.
Following the convention in meta-learning literatures~\cite{santoro2016meta,finn2017model_agnostic}, we call the tracking simulation as \textit{an episode} or \textit{a simulated tracking episode}.

The first step of a simulated tracking episode is the initial adaptation; the model parameter is learned for an initial dataset $\datainit$ by optimizing Eq.~\eqref{eq:adaptation}, where $\datainit$ is collected based on the target ground-truth annotation in a frame sampled from $\mathcal{V}$.
The updated model parameter is then employed to estimate the target states in the unseen frames sampled from $\mathcal{V}$.
The estimated targets are used to construct an online dataset $\dataonline$ for online model adaptation.
The online adaptation simulates the model adaptation using noisy labels, which commonly happens in real object tracking scenarios.
Note that the optimization of Eq.~\eqref{eq:adaptation} during the initial and online adaptations is guided by the hyperparameters that we learn in the meta-learning.
The final step is to collect a test dataset $\dataval$ for meta-training to define a test loss $\lossval$ and a meta-optimization loss $\lossmeta$, which are the objective functions of our meta-learning.
Note that $\dataval$ is obtained from the ground-truth annotations in $\mathcal{V}$ and augmented with the negative samples collected from all other annotated videos in the training video set $\Gamma - \{\mathcal{V}\}$.
Figure~\ref{fig:overall_pipeline} illustrates the overall pipeline in a simulated tracking episode.

A simulated tracking episode results in a set of the collected datasets and a series of intermediate model parameters associated with the episode.
Such information---datasets and parameters---is referred to as \textit{an episode trajectory} and denoted by $\tau$\footnote{The formal definition of $\tau$ can be found in ``Meta-parameter optimization'' subsection in page 4.}.
The objective function for meta-learning is defined over the trajectories from multiple episodes, where each episode is based on an annotated video $\mathcal{V} \in \Gamma$.
We optimize the hyperparameters based on this objective function.
Algorithm~\ref{pre_train_algorithm} describes the meta-learning procedure with the proposed simulated tracking episodes.

\begin{algorithm}[t]
	\caption{Meta-Learning for Fast Adaptation}\label{pre_train_algorithm}
	\begin{algorithmic}[1]
		\State
		\textbf{Input:} A Training video set $\Gamma$,
		\newline
		Meta-parameters $\mathcal{M}=\{\thetainit, \Ainit, \Aonline\}$,
		\newline
		\textbf{Output:}  Learned meta-parameters $\mathcal{M}^*$  \newline
		\While{not converged}
			\State{Sample a mini-batch of annotated videos from $\Gamma$}
			\For{\textbf{all} video $\mathcal{V} $ in a mini-batch}
				\State{Collect $\datainit$ based on $\mathcal{V}$}
				\For{\textbf{all} $k$ in $1,...,\Kinit$} ~// Eq.~\eqref{eq:strong_adaptation}
				\State{$\thetainitadapt^k = \thetainitadapt^{k-1} - \alphainit^k \odot \nabla_{\thetainitadapt^{k-1}}\mathcal{L}(\datainit; \thetainitadapt^{k-1})$}
				\EndFor{\textbf{end for}}
				\State{Collect $\dataonline$  based on $\mathcal{V}$ and $\thetainitadapt^{\Kinit}$}
				\For{\textbf{all} $k \in {1,...,\Konline}$}  ~// Eq.~\eqref{eq:weak_adaptation}
				\State{$\thetaonlineadapt^k = \thetaonlineadapt^{k-1} - \alphaonline^k \odot \nabla_{\thetaonlineadapt^{k-1}}\mathcal{L}( \dataonline; \thetaonlineadapt^{k-1})$}
				\EndFor{\textbf{end for}} ~// where $\thetaonlineadapt^{0} = \thetainitadapt^{\Kinit}$
				\State{Collect $\dataval$ based on $\mathcal{V}$ and $\Gamma - \{\mathcal{V}\}$}
				\State{Set $\tau=(\datainit, \dataonline, \dataval, \{\thetainitadapt^k\}_{k=1}^{\Kinit}, \{\thetaonlineadapt^k\}_{k=1}^{\Konline})$}
			\EndFor{\textbf{end for}}
			\State{Compute $\lossmeta$ over a mini-batch of $\tau$~// Eq.~\eqref{eq:meta_validation}}
			\State{$\mathcal{M} \leftarrow \text{optimize}(\mathcal{M}, \nabla_\mathcal{M}\mathcal{L}_\text{meta})$}  ~// Eq.~\eqref{eq:meta_parameters_update}
		\EndWhile{\textbf{end while}}
	\end{algorithmic}
\end{algorithm}
%

\subsubsection{Meta-parameters}

The meta-parameter $\mathcal{M}$ is a set of hyperparameters that are optimized by our meta-learning approach.
The primary objective of meta-learning is to reduce computational cost and maintain accuracy, and $\mathcal{M}$ allows to find a better trade-off between computation and performance by making a sophisticated control of a gradient-based model adaptation procedure.
We regard the initial model parameter $\thetainit \in \mathbb{R}^d$ as a hyperparameter.
In addition, we include learning rates as hyperparameters by extending a conventional scalar learning rate $\alpha$ to a per-parameter and per-iteration learning rate vector for the initial adaptation $\Ainit=\{\alphainit^k \in \mathbb{R}^d\}^{\Kinit}_{k=1}$ and the online adaptation $\Aonline=\{\alphaonline^k \in \mathbb{R}^d\}^{\Konline}_{k=1}$, where $\Kinit$ and $\Konline$ are the number of gradient-descent iterations for the initial and online adaptations, respectively.
The meta-parameter is a collection of all the hyperparameters, $\mathcal{M}=\{\thetainit,\Ainit,\Aonline\}$.

\subsubsection{Initial adaptation} 
The initial adaptation uses the initial dataset $\datainit$, which is constructed based on a frame and its ground-truth annotation that are uniformly sampled from an annotated video $\mathcal{V}$;
$\datainit$ contains positive samples that are tightly overlapped with the ground-truth annotation and negative samples that are loosely overlapped or separated from the annotation.
The procedure for sampling the positive and negative examples is identical to the strategy of dataset collection for target appearance learning in the standard object tracking~\cite{jung2018realtime}.

We adapt the initial model parameter $\thetainit$ by the stochastic gradient-descent method for $\Kinit$ iterations using Eq.~\eqref{eq:adaptation}, which is identical to real tracking scenarios.
As a result of the model update, we obtain a target appearance model learned from a ground-truth annotation in a single frame.
An iteration of the initial adaptation is given by
\begin{equation}
\thetainitadapt^k = \thetainitadapt^{k-1} - \alphainit^k \odot \nabla_{\thetainitadapt^{k-1}}\mathcal{L}(\datainit ; \thetainitadapt^{k-1}),
\label{eq:strong_adaptation}
\end{equation}
where  $k=1, \dots, \Kinit$.
The model parameter after the initial adaptation is given by $\thetainitadapt^{\Kinit}$.

\paragraph{Online Adaptation}
The online adaptation uses the online dataset $\dataonline$, which includes the examples sampled from the estimated targets.
The model parameterized by $\thetainitadapt^{\Kinit}$ is adopted to estimate the target states in the sampled frames by scoring candidate image patches.
Similarly to the standard object tracking, the candidate image patches are sampled from a local area centered at the ground-truth target, and an image patch with a maximum score in each frame is selected as an estimated target.
The dataset $\dataonline$ contains positive and negative samples collected based on the estimated targets; the data collection strategy is identical to the one for $\datainit$, except that the estimated targets are used.
Note that using the estimated targets for online adaptation turns out to be effective in handling the potential issues given by unreliable target estimation during tracking.

For online adaptation, we update $\thetainitadapt^{\Kinit}$ by stochastic gradient descent for $\Konline$ iterations based on Eq.~\eqref{eq:adaptation}.
The formal description of the online adaptation is given by 
%
\begin{equation}
\thetaonlineadapt^k = \thetaonlineadapt^{k-1} - \alphaonline^k \odot \nabla_{\thetaonlineadapt^{k-1}}\mathcal{L}(\dataonline; \thetaonlineadapt^{k-1}),
\label{eq:weak_adaptation}
\end{equation}
where $\thetaonlineadapt^0 = \thetainitadapt^{\Kinit}$ and $k =1, \dots ,\Konline$.
The model parameter obtained from the online adaptation is denoted by $\thetaonlineadapt^{\Konline}$.

\subsection{Hard example mining for meta-training}

Meta-learning aims to learn the meta-parameters maximizing the expected tracking accuracy of adapted models based on diverse tracking scenarios.
In meta-training, we thus evaluate a trained model by testing its tracking performance. 
As explored in Meta-Tracker~\cite{park2018metatracker}, the most straightforward choice for evaluating the model parameter is to measure its accuracy based on the ground-truth target annotation in the subsequent frames.
However, this approach might not be sufficient for evaluating the robustness of the model because the dataset typically has limited diversity.
Hence, we propose to use hard example mining to evaluate the model during testing in meta-training.

\subsubsection{Hard example dataset}

The test dataset in meta-training is a union of two datasets as $\dataval=\dataval^\text{std}\cup\dataval^\text{hard}$, where $\dataval^\text{std}$ is obtained from $\mathcal{V}$ and $\dataval^\text{hard}$ is collected from $\Gamma-\{\mathcal{V}\}$.
$\dataval^\text{std}$ consists of both positive and negative samples for a target in $\mathcal{V}$ and the strategy for collecting $\dataval^\text{std}$ is identical to the one for collecting $\datainit$, including the use of the ground-truth annotation.
$\dataval^\text{std}$ is essential for evaluating learned models, but it may not be sufficient to handle diverse distracting objects.
To consider such objects, we introduce $\dataval^\text{hard}$, which consists of the image patches obtained from the ground-truth annotations sampled from $\Gamma-\{\mathcal{V}\}$.
We label all examples in $\dataval^\text{hard}$ as negative.
Since these negative samples are actually target instances in the other videos, they are hard examples.

\subsubsection{Test objective in meta-training}

To test the model with the adapted model parameters $\thetageneral$ in meta-training, we compute a test loss on the test dataset $\dataval$, which is given by a combination of the cross-entropy loss in Eq.~\eqref{eq:adaptation} and the triplet loss defined below:
\begin{equation}
\lossval(\dataval;\thetageneral) = \mathcal{L}(\dataval^\text{std};\thetageneral) +\gamma \losstriple(\dataval;\thetageneral)
\end{equation}
where $\gamma$ is a hyperparameter controlling the relative importance of the two terms.
The triplet loss is defined by
\begin{align}
&\losstriple (\dataval; \thetageneral)= \\
&\hspace{0.4cm}\mathbb{E}_{p_{\dataval}(\x, \x^{+}, \x^{-})}\Big[ \big(\xi + \Delta(\x, \x^{+}; \thetageneral) - \Delta(\x, \x^{-}; \thetageneral)\big)_+ \Big], \nonumber
\end{align}
where $p_{\dataval}(\x, \x^+, \x^-)$ denotes a distribution of triplets, $(\x, \x^+, \x^-)$, determined by the sampling strategy from  $\dataval$, $(\cdot)_+ \eqdef \max(0, \cdot)$, and $\xi$ is a margin. 
Also, $\Delta(\x_1, \x_2; \thetageneral)$ means the Euclidean distance between $L_2$ normalized outputs of the model as follows:
\begin{align}
\Delta(\x_1, \x_2; \thetageneral) \defeq \left|\left| \frac{\f(\x_1 ;\thetageneral)}{\|\f(\x_1;\thetageneral) \|^2} - \frac{\f(\x_2;\thetageneral)}{\|\f(\x_2;\thetageneral)\|^2} \right|\right|^2.
\end{align}

Note that hard examples are involved for the computation of the triplet loss because we draw two positive samples $(\x, \x^+)\sim\dataval^\text{std}$ and a hard examples $\x^- \sim \dataval^\text{hard}$.
Since the hard examples can be positive in some videos, we have to distinguish the hard examples from pure background instances.
The cross-entropy loss cannot achieve this goal, and this is why we introduce the triplet loss and make the positive examples from a video clustered together while embedding them to be separated from the positives in the rest of videos  and the backgrounds.

\subsection{Meta-parameter optimization}

We optimize the meta-parameters to encourage the adapted model parameters to minimize the expected meta-optimization loss, which is defined over multiple simulated tracking episodes.
Given an episode trajectory of a tracking simulation, $\tau=(\datainit, \dataonline, \dataval, \{\thetainitadapt^k\}_{k=1}^{\Kinit}, \{\thetaonlineadapt^k\}_{k=1}^{\Konline})$, which is a collection of all the datasets and intermediate model parameters used in the episode, we optimize the meta-parameters using the following meta-optimization loss
\begin{align}
\label{eq:meta_validation}
\lossmeta&(\mathcal{M}) = \\
&\mathbb{E}_{p(\tau)}\big[\lossval(\dataval;\thetainitadapt^{\Kinit}) + \lossval(\dataval;\thetaonlineadapt^{\Konline})\big], \nonumber
\end{align}
where $p(\tau)$ is a distribution of episode trajectories, defined by the strategy for sampling a video $\mathcal{V}$ from a video set $\Gamma$ and for performing tracking simulation with meta-parameter $\mathcal{M}$.
As shown in Figure~\ref{fig:overall_pipeline} and Eq.~\eqref{eq:meta_validation}, the meta-optimization loss adopts the test loss whose parameters $\thetainitadapt^{\Kinit}$ and $\thetaonlineadapt^{\Konline}$ are further represented by the meta-parameters $\mathcal{M}$.
It implies that our meta-optimization loss $\lossmeta$ is fully differentiable with respect to the meta-parameters $\mathcal{M}$ for optimization.
%
%
By employing an optimizer such as ADAM~\cite{diederik2014adam}, we update the meta-parameters using the following gradients:
\begin{equation}
\begin{split}
&
\nabla_\mathcal{M}\lossmeta=\mathbb{E}_{p(\tau)}\bigg[
\frac{\partial \lossval(\dataval;\thetainitadapt^{\Kinit})}{\partial\thetainitadapt^{\Kinit}}
\frac{\partial\thetainitadapt^{\Kinit}}{\partial\mathcal{M}}
\\
&\hspace{0.5cm}
+
\frac{\partial \lossval(\dataval;\thetaonlineadapt^{\Konline})}{\partial\thetaonlineadapt^{\Konline}}
\bigg\{
\frac{\partial\thetaonlineadapt^{\Konline}}{\partial\mathcal{M}}
+
\frac{\partial\thetaonlineadapt^{\Konline}}{\partial\thetainitadapt^{\Kinit}}
\frac{\partial\thetainitadapt^{\Kinit}}{\partial\mathcal{M}}
\bigg\}
\bigg]
\label{eq:meta_parameters_update}
\end{split}
\end{equation}
Eq.~\eqref{eq:meta_parameters_update} illustrates that the meta-parameters are updated by optimizing tracking results, which implies that the meta-parameters $\mathcal{M}$ induce a strong object tracker based on the proposed fast adaptation algorithm.

\section{Meta-learning One-Shot Channel Pruning}

We present how to meta-learn to remove redundant channels given a ground-truth target bounding box at the first frame of a video.
Our pruning network builds on the LASSO regression-based channel pruning~\cite{he2017channel} and extends it by learning to predict a channel selection mask for an entire tracking episode from the first frame of a video. 
As in Section~\ref{sec:meta_learn_fast_adaptation}, our meta-learning for pruning is trained using simulated tracking episodes and thus enables the network to predict a mask appropriate for the future frames even in a previously unseen video. 
We first formulate the channel pruning task in a tracking episode and discuss how to leverage meta-learning for one-shot channel pruning.
In this section, for notational simplicity, we regard a fully-connected layer as a $1 \times 1$ convolution layer.

\subsection{Learning channel masks in a tracking simulation}
We build on the LASSO pruning~\cite{he2017channel} that learns channel selection masks $\mathcal{B}=\{\boldsymbol{\beta}^l\}_{l=1}^{L}$, where $L$ is the number of layers in a CNN.
The pruning network is trained to predict mask $\mathcal{B}$, which is regularized by sparsity constraints, so that the selected maps by mask $\mathcal{B}$ approximate original feature maps of the CNN.  
A channel selection mask $\boldsymbol{\beta}^l$ is a real-valued vector whose dimension is identical to the number of the channels in the $l^\text{th}$ layer.
We prune the channels corresponding to the zero-valued dimensions in $\boldsymbol{\beta}^l$.

Our LASSO pruning minimizes the following loss function:
\begin{equation}
\begin{split}
\textstyle \losslasso&(\mathcal{D}, \mathcal{B} ; \thetageneral)
	= \textstyle  \lambda \sum_{l=1}^L\big\| \boldsymbol{\beta}^l\big\|_1
	\\
	+ & \textstyle \mathbb{E}_{p_{\mathcal{D}}(\x)} \Big[
		\sum_{l\in\mathcal{S}} 
			\big\|F^l(\x; \boldsymbol{\theta}) -  F^l_\mathcal{B}(\x; \boldsymbol{\theta})\big\|_2  \Big]  ,
\label{eq:lasso_prune}
\end{split}
\end{equation}
%
where $\lambda$ controls the degree of sparsity in channel selection and $\mathcal{S}$ is a set of layers to approximate the original feature map.
$F^l(\x ; \boldsymbol{\theta})$ denotes a feature map in the $l^\text{th}$ layer while $F^l_ \mathcal{B}(\x; \boldsymbol{\theta})$ represents a feature map pruned by mask $\mathcal{B}$. 

Given an input image patch $F^0(\x;\thetageneral)=F^0_{\mathcal{B}}(\x;\thetageneral) = \x$, unmasked network sequentially computes features by $F^{l+1}(\x; \boldsymbol{\theta})=\sigma\big(\text{Conv}(F^{l}(\x; \boldsymbol{\theta}); \boldsymbol{\theta}^l ) \big)$, where the parameters for $l^\text{th}$ layer $\boldsymbol{\theta}^l$ and $\sigma(\cdot)$ is a non-linear activation function. 
Here we omit some components including pooling layers for notational simplicity. 
In contrast, the pruned feature maps are computed along layers as 
\begin{equation}
F^{l+1}_\mathcal{B}(\x ; \boldsymbol{\theta}) = \boldsymbol{\beta}^{l+1} \odot \sigma\big( \text{Conv} (F^l_\mathcal{B}(\x ; \boldsymbol{\theta}); \boldsymbol{\theta}^l)  \big). 
\end{equation}



To consider pruning for a simulated tracking episode, we extend the objective of Eq.~\eqref{eq:lasso_prune}, defined on a dataset $\mathcal{D}$, into the following objective for an episode trajectory $\tau$:  
\begin{equation}
\begin{split}
& \textstyle \lossepisode(\tau, \mathcal{B})
	 = \losslasso(\datainit, \mathcal{B}; \thetainitadapt^{\Kinit}) \\
& \textstyle \hspace{0.2cm}
	+ \sum_{k=1}^{\Konline}\losslasso(\dataonline, \mathcal{B};\thetaonlineadapt^k)
	+ \losslasso(\dataval, \mathcal{B};\thetaonlineadapt^{\Konline}). 
\label{eq:lasso_prune_episode}
\end{split}
\end{equation}
This objective aims to find a single $\mathcal{B}$ that minimizes $\losslasso$ over different pairs of parameters and datasets appearing in a single simulated tracking episode.

\begin{table*}[t!]
	\begin{center}
		\caption{
			Analysis of proposed components for meta-learning on OTB2015.
		}
		\label{tab:ablation}
		\resizebox{2.1\columnwidth}{!}{
			\begin{tabular}{
					l|cccccc|ccc
			}
				Method & $\dataonline$ & $\dataval^\text{hard}$ & $\thetainit$  & $\Ainit$ &  $\Aonline$ &  Prunning  & Succ(\%) & Prec(\%) & FPS\\ 
				\hline
				\hline
				RT-MDNet~\cite{jung2018realtime} & {-}    & {-} & {-}  &  {-}     & {-}   &{-}   &   65.0     & 88.5 & 42 \\
				\hline
				MetaRTT        & $\surd$& $\surd$ & $\surd$  & $\surd$ & $\surd$ & {} & 65.5 &  89.0 & 58\\
				~~~-~Not using estimated target       & {} & $\surd$  & $\surd$ & $\surd$ & $\surd$ &{}&      58.6     & 81.2 & 54 \\
				~~~-~Not using hard examples         & $\surd$ & {} & $\surd$ &   $\surd$   &  $\surd$ &    & 60.7  & 81.9 & 57 \\
				~~~-~Not using per-parameter / per-iteration lr (scalar lr)      & $\surd$& $\surd$ & $\surd$  & {} & {} &       &  60.9  & 82.8&  57\\
				\hline
				MetaRTT+Prune  & $\surd$& $\surd$ & $\surd$  & $\surd$ & $\surd$ & $\surd$  & 65.1 & 87.4 & 65 \\
				\hline
				Baseline for Meta-Tracker~\cite{park2018metatracker} &{}&  {} & $\surd$  & $\surd$    & {}   &{}     &   56.8 & 81.2 & 40 \\
				~~~-~Using hard examples &{} & $\surd$ & $\surd$   & $\surd$  & {}  & {}   & 62.7      & 86.2     & 41\\
				\hline
			\end{tabular}
		}
	\end{center}
\end{table*}

\subsection{Meta-learning for channel mask prediction}

The objective of Eq.~\eqref{eq:lasso_prune_episode} uses an episode trajectory from an entire simulated tracking episode to learn channel selection masks.
However, speeding up the object tracking requires channels to be pruned even before the complete trajectory for a tracking episode is collected.
Therefore, we further introduce one-shot channel pruning that amortizes optimization in Eq.~\eqref{eq:lasso_prune_episode} as a prediction of $\mathcal{B}$ based on a dataset coming from an initial frame of a video.


We build our pruning network on top of the CNN used for object tracking.
It predicts $\mathcal{B}$ using $\datainit$ and $\thetageneral$.
The prediction is denoted by a function outputting a set of vectors $\Psi(\datainit; \thetageneral, \phigeneral) = \{\psi^l(\datainit; \thetageneral,\phigeneral)\}_{l=1}^{L}$, where $\psi^l(\datainit; \thetageneral, \phigeneral)$ approximates $\boldsymbol{\beta}^l$, a channel mask for the $l^\text{th}$ layer of the CNN.
Note that $\phigeneral \in \mathbb{R}^c$ is a parameter of the pruning network that is not shared with the tracking CNN.
Specifically, we construct the pruning network as $\psi^l(\datainit; \thetageneral, \phigeneral) = \frac{1}{N}\sum_{i=1}^{N} \text{MLP}\big(\text{AvgPool}\big(F^l(\x_i; \boldsymbol{\theta})\big);\phigeneral^l\big)$, where
$\text{MLP}(\cdot;\phigeneral^l)$ is a multi-layer perceptron parameterized by $\phigeneral^l$ associated with the $l^\text{th}$ layer, $\text{AvgPool}(\cdot)$ is a spatial global average pooling, and $N$ is the number of samples in $\datainit$.
Note that we pretrain $\boldsymbol{\theta}$ via meta-learning for fast adaptation and keep it fixed while optimizing $\phigeneral$ for training the pruning network.

The meta-learning objective of our one-shot pruning network $\Psi(\datainit;\thetageneral, \phigeneral)$ is to minimize the loss
\begin{align}
\lossprune(\phigeneral) = \mathbb{E}_{p(\tau)} \Big[
	\lossepisode\big(\tau, \Psi(\datainit; \thetainitadapt^{\Kinit}, \phigeneral)\big)
\Big],
\label{eq:lasso_prune_meta}
\end{align}
which substitute $\mathcal{B}$ in Eq.~\eqref{eq:lasso_prune_episode} with $\Psi(\datainit; \thetainitadapt^{\Kinit}, \phigeneral)$ and train it with the samples from trajectory distribution $p(\tau)$, which is identical to the distribution in Eq.~\eqref{eq:meta_validation}. 
In this work, we use $\thetainitadapt^{\Kinit}$ in channel mask prediction $\Psi(\datainit; \thetainitadapt^{\Kinit}, \phigeneral)$ so that the pruning network can exploit the parameters specialized for a video using the first frame with a target annotation.
Since $\thetainitadapt^{\Kinit}$ is obtained from the initial frame of the video, our pruning network does not incur an additional computation cost in prediction at the future frames. Eq.~\eqref{eq:lasso_prune_meta} is a meta-learning objective as it defines an outer learning objective for an inner learning objective of Eq.~\eqref{eq:lasso_prune_episode}.

\newcommand{\rowgroup}[1]{\hspace{-0em}#1}

\section{Experiments}
\label{sec:experiments}

This section discusses the details of our meta-learning applied to RT-MDNet~\cite{jung2018realtime} and the performance of the proposed algorithm in comparison to the state-of-the-art methods.
We present the results from three versions of our trackers; MetaRTT is the model without one-shot network pruning, MetaRTT+Prune is the one with pruning, and MetaRTT+COCO is the model trained with additional data. 
The base model for all trackers is RT-MDNet.

\subsection{Application to RT-MDNet}

We apply our meta-learning algorithm to RT-MDNet, which is one of the state-of-the-art real-time object trackers.
For efficient feature extraction from multiple image patches in a frame, RT-MDNet employs a two-stage feature extraction pipeline similar to \cite{he2017maskrcnn}.
In the first stage, a convolutional neural network (CNN) computes a feature map from an input frame.
To extract the feature descriptor from the shared feature map corresponding to an image patch, RT-MDNet employs RoIAlign layer~\cite{he2017maskrcnn}, where a $3 \times 3$ dimensional feature map grid is interpolated and pooled.
The feature descriptor for an image patch $\x$ is employed to compute the response for the patch using three fully connected layers, which is denoted by $\f(\x; \boldsymbol{\theta})$.

Our definition of parameters $\thetainit$ includes the parameters both before and after the RoIAlign layer.
Therefore, the initial parameter $\thetainit$ on both sides are meta-learned.
However, similarly to the original RT-MDNet, the initial and online adaptations are not applied to the parameters for the shared CNN both during the tracking and the tracking simulation.
Formally, there is no model adaptations in the shared CNN layers, {\it i.e.,} $\thetainitadapt^k=\thetainitadapt^{k-1}$ and $\thetaonlineadapt^k=\thetaonlineadapt^{k-1}$, while the model adaptions for the fully connected layers are given by Eq.~\eqref{eq:strong_adaptation} and~\eqref{eq:weak_adaptation}.
For meta-learning one-shot channel pruning, we define $\mathcal{S}$ in Eq.~\eqref{eq:lasso_prune} as a set of the final shared convolutional layer and all the fully connected layers.
We set $\Kinit$ and $\Konline$ to 5 throughout our experiment.

\subsection{Implementation details}
We pretrain MetaRTT and MetaRTT+Prune on ImageNet-Vid~\cite{russakovsky2015imagenet}, which contains more than 3,000 videos with 30 object classes labeled for video object detection. 
We treat each object as a class-agnostic target to track but do not use its class label for pretraining.
We randomly select 6 frames from a single video to construct an episode, and use the first frame for $\datainit$, the last frame for $\dataval^\mathrm{std}$ and the remaining frames for $\dataonline$.
The meta-parameters are optimized over 40K simulated episodes using ADAM with fixed learning rate $10^{-4}$.
The network learning the pruning mask per layer is implemented as a 2-layer perceptron with a LeakyRelu activation and dropout regularization. 
We optimize the network by ADAM for 30K iterations with learning rate $5 \times 10^{-5}$. 
Other details follow the configuration of RT-MDNet. 
We pretrain MetaRTT+COCO on ImageNet-Vid and the augmented version of COCO~\cite{lin2014microsoft}.
Since COCO is a dataset of images, we construct its augmented set by artificially generating pseudo-videos with 6 frames via random resizing and shift of the original images as in \cite{zheng2018distractor}.
Our algorithm is implemented in PyTorch with 3.60 GHz Intel Core I7-6850k and NVIDIA Titan Xp Pascal GPU.
\subsection{Ablation studies}

\subsubsection{Meta-learning for fast adaptation}
Table~\ref{tab:ablation} shows ablative results on OTB2015~\cite{wu2013online}, where we show the performance of several different model variants given by the adopted algorithm components.
Specifically, $\dataonline$ denotes whether each option employs the online adaptation based on the estimated targets; if the corresponding column is not checked, it means that meta-learning is performed using the ground-truth labels only.
Also, $\dataval^\text{hard}$ indicates the use of hard examples for meta-training.

As expected, individual algorithm components make substantial contribution to improving accuracy and speed of our tracker.
Our full algorithm maintains competitive accuracy and achieves significant speed-up compared to the baseline technique, RT-MDNet.
For comparison with other meta-learning approaches for object tracking, we select MetaTracker~\cite{park2018metatracker}. 
Unlike our meta-learning framework, Meta-Tracker employs meta-learning only for the initial adaptation, and it does not utilize the estimated targets for the online adaptation and the hard example mining for meta-training. 
According to our experiment, the meta-learning strategy proposed by Meta-Tracker shows relatively poor performance compared to MetaRTT while we also observe that the success rate and the precision of Meta-Tracker are improved by simply incorprating the component exploiting hard examples.

Table~\ref{tab:fast_adapt} presents how the number of iterations affects performance of our algorithm with meta-learning, MetaRTT, in comparison to RT-MDNet.
MetaRTT achieves better success rate and precision while it has 38\% of speed-up compared to RT-MDNet through a more efficient model adaptation.
Note that a simple reduction of model update iterations in RT-MDNet degrades its accuracy while increasing speed.

\begin{table}[t!]
	\begin{center}
		\caption{
			Accuracy comparison with respect to the number of iterations $\Kinit$ and $\Konline$ for fast adaptation on OTB2015.
		}
		\label{tab:fast_adapt}
		\resizebox{0.9\columnwidth}{!}{
			\begin{tabular}{
					l|c|c|ccc
				}
				Method & $\Kinit$ & $\Konline$ &  Succ (\%) & Prec (\%) & FPS\\ 
				\hline
				\hline
				\multirow{4}{*}{RT-MDNet} & 50 & 15       &  65.0 & 88.5 & 42 \\
				&50 & 5&  61.9 & 83.1 & 54\\
				& 5 & 15       &  60.6  & 80.8 & 43\\
				& 5 & 5     &   57.8 & 77.5& 55\\
				\hline
				MetaRTT  & 5 & 5 &  65.5 & 89.0 & 58\\
				\hline
			\end{tabular}
		}
	\end{center}
\end{table}

\begin{table}[t!]
	\begin{center}
		\caption{
			Analysis of one-shot channel pruning on OTB2015. PR denotes prune rate in this table.
		}
		\label{tab:ablation_prune}
		\resizebox{0.99\columnwidth}{!}{
			\begin{tabular}{l|cccc}
				Method &  Succ (\%) & Prec (\%) & FPS & \makecell{PR (\%)}\\ 
				\hline
				\hline
				MetaRTT               & 65.5 & 89.0 & 58 & 0   \\
				\hline
				MetaRTT+Prune      & 65.1 & 87.4 &  65 & $50\pm{5}$ \\
				-~No meta-learning     & 61.7 & 83.1 & 64 & $49\pm{6}$  \\
				-~No adaptive pruning     & 60.9 & 81.2 & 59 & 53              \\
				\hline
			\end{tabular}
		}
	\end{center}
\end{table}

\subsubsection{Meta-learning for one-shot channel pruning}
Table~\ref{tab:ablation} presents that MetaRTT+Prune is $12$\% faster than MetaRTT while achieving the almost identical success rate and precision. 
To analyze the proposed pruning method, we evaluate two variants of pruning strategies.
One is the one-shot channel pruning without meta-learning, which optimizes the pruning network using a simulated tracking episode only at the first frame.
The other option is the static pruning, which learns a universal channel pruning mask $\mathcal{B}$ offline for every video.
Table~\ref{tab:ablation_prune} implies that both the meta-learning with complete simulated episodes and the input adaptive one-shot pruning are essential.
It also presents that MetaRTT+Pruning improves speed substantially by removing almost half of the network parameters with minimal accuracy loss.

\subsection{Comparison with state-of-the-art trackers}
We compare variants of our meta-learning algorithm---MetaRTT, MetaRTT+Prune, and MetaRTT+COCO---to recent state-of-the-art algorithms~\cite{nam2016learningmulti,martin2017eco,martin2016beyond,martin2017discriminative,fan2017parallel,jung2018realtime,zheng2018distractor}.
We employ OTB2015~\cite{wu2015object} and TempleColor~\cite{liang2015encoding} datasets for evaluation.
Because our goal is to accelerate tracking algorithms while maintaining their accuracy, the comparison with real-time trackers is more important.
So, we highlight real-time trackers with solid lines in the graphs visualizing quantitative performance while the slow methods are represented with dashed lines.

\begin{figure}[t!]
\centering
\includegraphics[width=0.49\columnwidth]{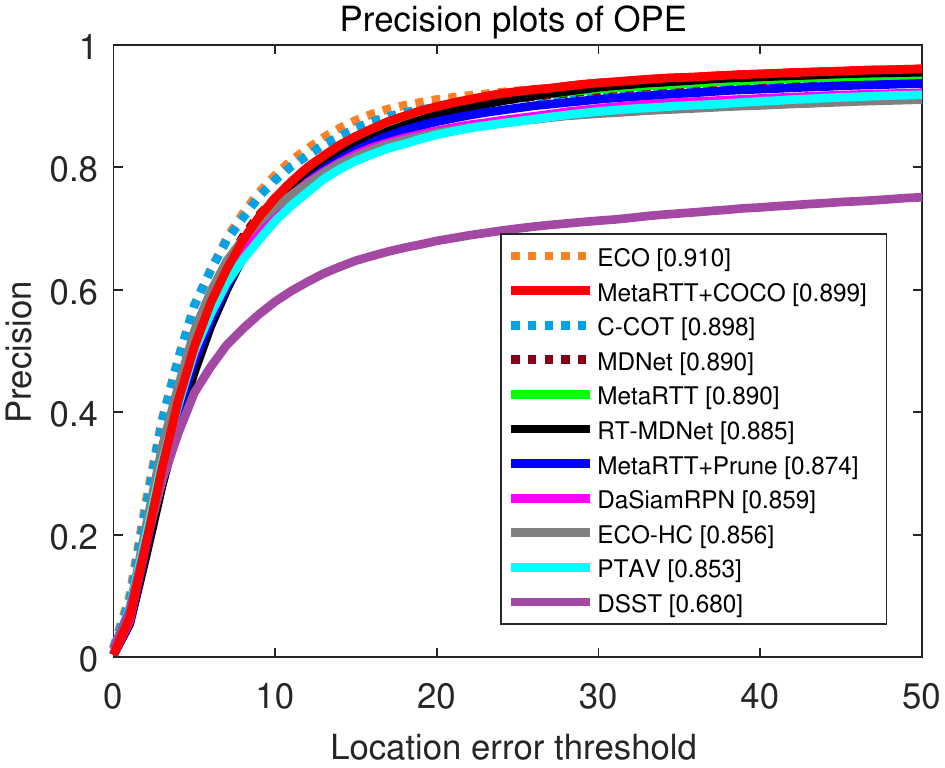}
\includegraphics[width=0.49\columnwidth]{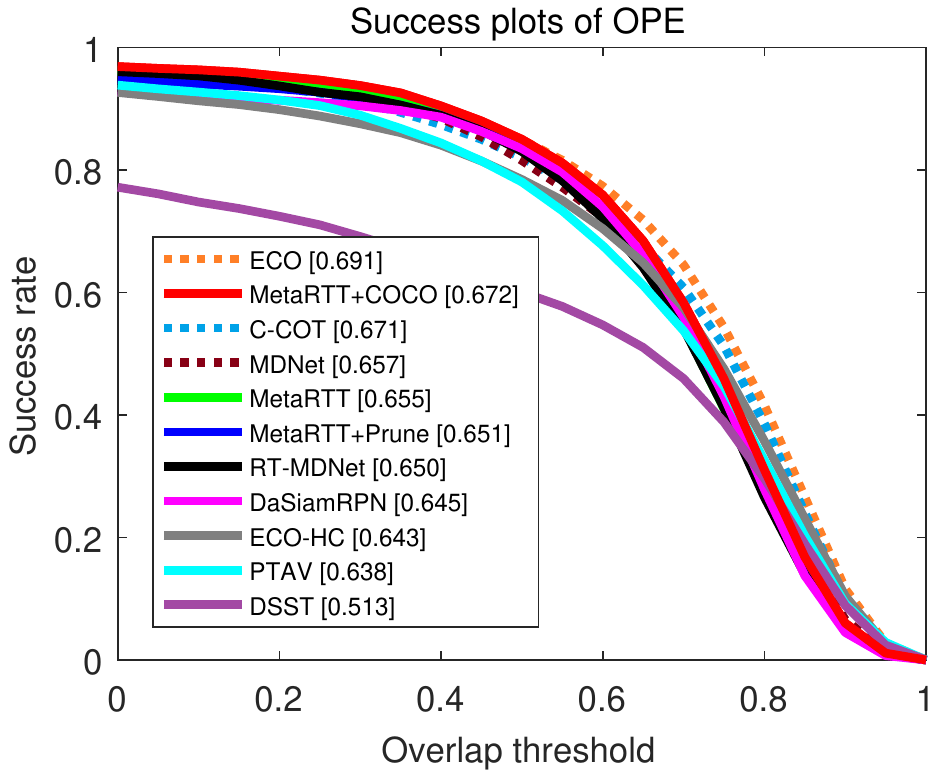}
\caption{Quantitative results on OTB2015.}
\label{fig:otb_plot}
\end{figure}

\begin{figure}[t!]
	\begin{center}
	\includegraphics[width=0.49\columnwidth]{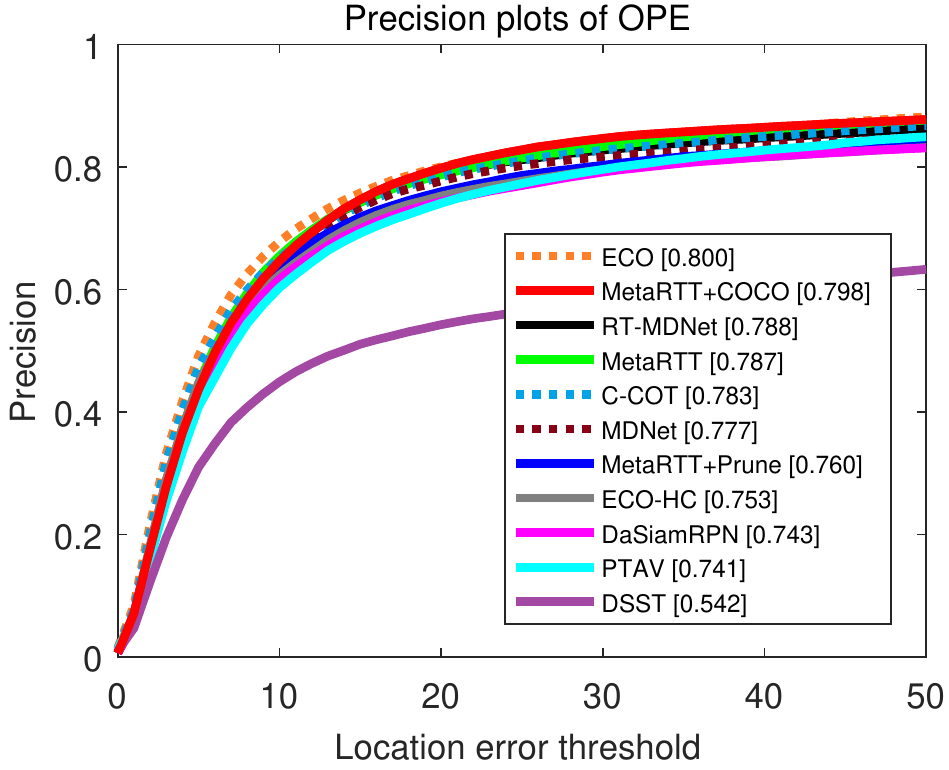}
	\includegraphics[width=0.49\columnwidth]{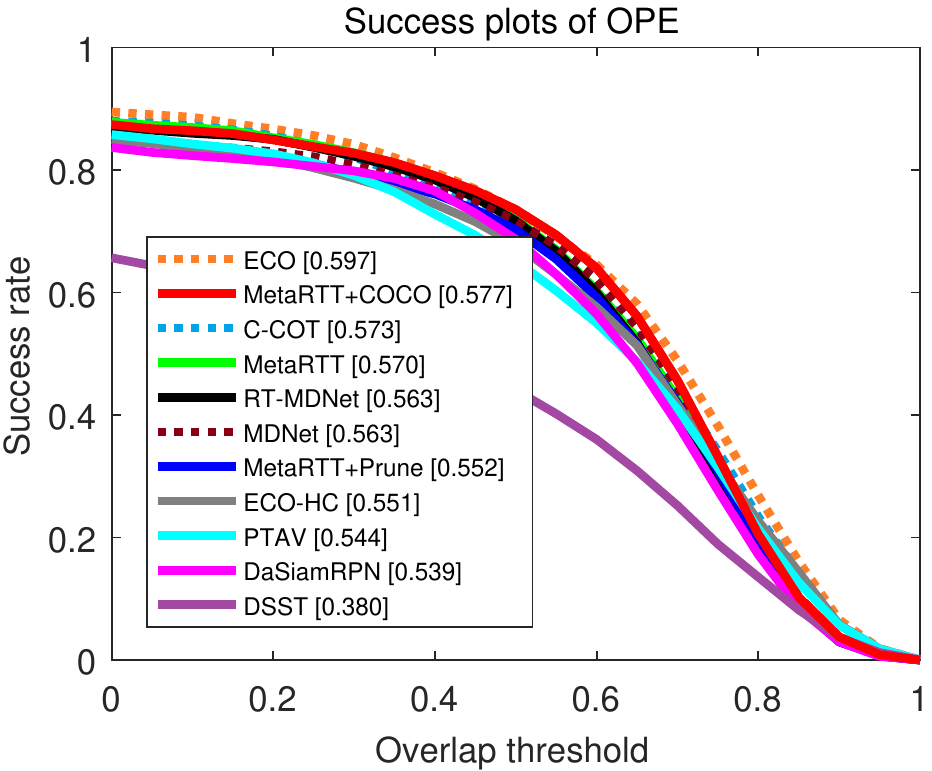}
	\end{center}
	\caption{Quantitative results on TempleColor.}
	\label{fig:temple_plot}
\end{figure}

Figure~\ref{fig:otb_plot} shows that both \textrm{MetaRTT} and \textrm{MetaRTT+COCO} outperform the compared state-of-the-art real-time trackers on OTB2015 in terms of accuracy while MetaRTT+Prune achieves competitive accuracy and improved speed compared to MetaRTT.
Figure~\ref{fig:temple_plot} shows that our trackers present outstanding performance on TempleColor as well. 
These results imply that the hyperparameters meta-learned on ImageNet-Vid performs well on the standard benchmarks without manual tuning for the benchmarks; we fix all the hyperparameters except the meta-parameters in both datasets.
The high accuracy by MetaRTT+COCO also suggests that our tracker can further improve by pretraining with more datasets.

Table~\ref{tab:ablation_votuav} presents that the variants of our algorithm make significant improvement on other benchmark datasets, VOT2016~\cite{kristan2016vot} and UAV123~\cite{matthias2016benchmark}.
\begin{table}[t!]
	\begin{center}
		\caption{
			Results of our algorithms on VOT2016, UAV123.	
		}
		\label{tab:ablation_votuav}
		\resizebox{0.9\columnwidth}{!}{
			\begin{tabular}{l|cc|c}
				    {}       &  \multicolumn{2}{c|}{UAV123} & VOT2016 \\ 
				Method &  Succ (\%) & Prec (\%) & EAO  \\ 
				\hline
				\hline
				RT-MDNet                   & 52.8 & 77.2 & 0.322 \\
				\hline
				MetaRTT                     & 56.9 & 80.9 & 0.346   \\
				MetaRTT+COCO  & 56.6 & 80.9 & 0.350   \\
				MetaRTT+Prune         & 53.7 & 77.9 &  0.314  \\
				\hline
			\end{tabular}
		}
	\end{center}
\end{table}

\section{Conclusion}
\label{sec:conclusion}
We have presented a novel meta-learning framework for object tracking, which is more comprehensive and principled than the technique introduced in previous studies.
The proposed meta-learning approach allows to learn fast adaptation and one-shot channel pruning algorithm, which leads to competitive accuracy and substantial speed-up.
Our tracker achieves the state-of-the-art performance compared to the existing real-time tracking methods.

\paragraph{Acknowledgement}
This work was partly supported by Institute for Information \& communications Technology Promotion(IITP) grant funded by the Korea government(MSIT)[2014-0-00059, 2017-0-01780, 2016-0-00563], LG-CNS Co., Ltd., Seoul, and NRF Korea (NRF-2017M3C4A7069369). 





{
\small
\bibliographystyle{aaai}
\bibliography{ref}
}

\end{document}